\documentclass[10pt,twocolumn,letterpaper]{article}

\usepackage{iccv}
\usepackage{times}
\usepackage{epsfig}
\usepackage{graphicx}
\usepackage{amsmath}
\usepackage{amssymb}
\usepackage{bbm}
\usepackage{amsfonts}
\usepackage{amsthm}
\usepackage{multirow}
\usepackage{enumitem}
\usepackage{framed}
\usepackage{mathtools} 
\usepackage{breqn}
\usepackage{footmisc}
\usepackage{makecell}
\usepackage{footnote}
\usepackage{algorithmic}
\usepackage{algorithm}
\makesavenoteenv{tabular}
\makesavenoteenv{table}

\usepackage[breaklinks=true,bookmarks=false]{hyperref}

\iccvfinalcopy 

\def\iccvPaperID{****} 

\ificcvfinal\fi

\begin{document}

\title{{Interpreting Attributions and Interactions of Adversarial Attacks}}

\author{Xin Wang$^{a*}$,  Shuyun Lin$^{a}$\thanks{{Equal contribution}}, Hao Zhang$^{a}$, Yufei Zhu$^{a}$, Quanshi Zhang$^{a}$\thanks{{Quanshi Zhang is the corresponding author. He is with the John Hopcroft Center and the MoE Key Lab of Artificial Intelligence, AI Institute, at the Shanghai Jiao Tong University, China.}} \\
$^{a}$Shanghai Jiao Tong University \\
}

\maketitle
\ificcvfinal\thispagestyle{empty}\fi

\begin{abstract}
This paper aims to explain adversarial attacks in terms of how adversarial perturbations contribute to the attacking task. We estimate attributions of different image regions to the decrease of the attacking cost based on the Shapley value. 
We define and quantify interactions among adversarial perturbation pixels, and decompose the entire perturbation map into relatively independent perturbation components. 
The decomposition of the perturbation map shows that adversarially-trained DNNs have more perturbation components in the foreground than normally-trained DNNs.
Moreover, compared to the normally-trained DNN, the adversarially-trained DNN have more components which mainly decrease the score of the true category.
Above analyses provide new insights into the understanding of adversarial attacks.
\end{abstract}

\section{Introduction}

Deep neural networks (DNNs) have shown promise in various tasks, such as image classification~\cite{imagenet2015} and speech recognition~\cite{speech2014}. 
Adversarial robustness of DNNs has received increasing attention in recent years. 
Previous studies mainly focused on attacking algorithms~\cite{intriguing2014,cw2017,pgd2018}, the detection of adversarial examples for the adversarial defense~\cite{Metzen2017detection,Feinman2017detecting,SafetyNet}, and adversarial training to learn DNNs robust to adversarial attacks~\cite{advtrain2,pgd2018}.

Unlike previous studies of designing more powerful attacks or learning more robust DNNs, in this research,  we aim to explain  the signal-processing behavior behind the adversarial attack, \emph{i.e.} how pixel-wise perturbations cooperate with each other to achieve the attack. 
We develop new methods to explain adversarial attacks from the following perspectives.

\begin{figure}[t]
  \centering
  \includegraphics[width=\linewidth]{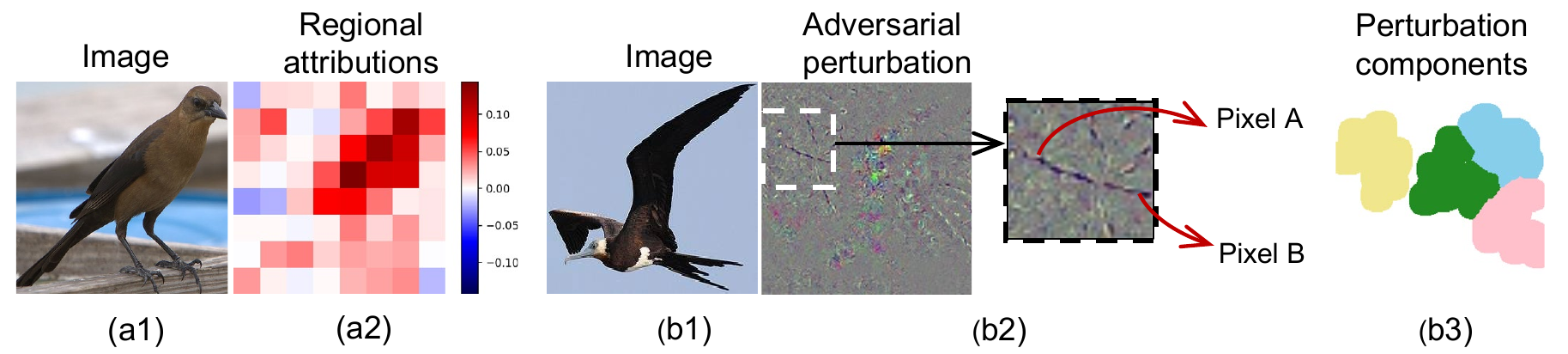}
  \vspace{-10pt}
  \caption{{(a) Regional attributions to the adversarial attack. Regions with high attributions are important for the decrease of the attacking cost. 
  (b2) Perturbation pixels A and B interact with each other and form a curve to conduct the adversarial attack; (b3) the entire perturbation can be decomposed into several components. Perturbation pixels within each component have strong interactions, whereas perturbation pixels between different components have relatively weak interactions.}} 
  \vspace{-10pt}
  \label{fig:1}
\end{figure}

{\textbf{1.} Given an input image,  \textbf{the regional attribution to the adversarial attack} is computed to diagnose the importance of each image region to the decrease of the attacking cost, \textit{i.e.} the $L_p$ norm of the adversarial perturbation.}
As Fig.~\ref{fig:1}~(a2) shows, regions of the bird's head and neck have high attributions to the adversarial attack.
If these two regions are not allowed to be perturbed, then magnitudes of adversarial perturbations in other regions would be significantly increased for attacking.
In this way, the attacking cost may increase significantly.

The regional attribution (importance) provides a new perspective to understand adversarial attacks.
We compute such regional attributions as Shapley values \emph{w.r.t.} the attacking cost.

\textbf{2. Pixel-wise interactions \& perturbation components in the adversarial attack:}
Given a perturbation map of the input image, we further define and quantify interactions among pixel-wise perturbations in the perturbation map, termed \textit{perturbation pixels}.
\emph{I.e.} we aim to explore how perturbation pixels cooperate to achieve the attack.
According to~\cite{Su2019Onepixel}, the adversarial power of a single pixel mainly depends on the context around the pixel, rather than rely on each perturbation pixel independently.
For instance, in Fig.~\ref{fig:1}~(b2), perturbation pixels $A$ and $B$ do not directly contribute to the attack. 
Instead, they interact with each other to form a curve to fool the DNN.

The interaction among perturbation pixels can be defined based on the game theory.
Given a DNN $g$ trained for classification and an adversarial image $x'=x+\delta\in\mathbb{R}^n$, $y=g(x')\in\mathbb{R}$ denotes the scalar output of the DNN (or one dimension of the vectorized network output).
{Let $\phi_i$ denote the importance (attribution) of the $i$-th perturbation pixel of $\delta$ \emph{w.r.t.} the output $y$, which is implemented as the Shapley value.
The attribution values of all perturbation pixels satisfy $g(x')-g(x)=\sum_{i=1}^n \phi_i$.}
Let $\phi_S$ denote the overall importance of all perturbation pixels in $S$, when perturbation pixels in $S$ collaborate with each other. 
Then, the interaction is defined as the change of the importance of $S$, when we ignore the collaboration between perturbation pixels and simply sum up the importance of each individual-working perturbation pixel in $S$, \emph{i.e.}  {\small $\phi_S-\sum_{i\in S}\phi_i$} quantifies the interaction.
If {\small $\phi_S-\sum_{i\in S}\phi_i>0$}, it indicates that perturbation pixels in $S$ cooperate with each other, and exhibit positive interactions. 
If {\small $\phi_S-\sum_{i\in S}\phi_i<0$}, it indicates that perturbation pixels in $S$ conflict with each other, and exhibit negative interactions.

Furthermore, based on the pixel-wise interactions among perturbation pixels,  as Fig.~\ref{fig:1}~(b3) shows, we can decompose the effect of the adversarial attack into several \textit{perturbation components}, which provides a new perspective to analyze how perturbation pixels cooperate with each other.
To this end, we develop a method to extract groups of perturbation pixels with strong interactions as perturbation components.
Perturbation pixels in the same component have strong interactions with each other.
Whereas, perturbation pixels in different components have relatively weak interactions.

\textbf{Using the Shapely value for explanation and its advantages:} We define the regional attribution and interactions among perturbation pixels based on the Shapley value~\cite{shapley1953value}. Though explanation methods in previous studies, such as Grad-CAM~\cite{selvaraju2017grad} and GBP~\cite{springenberg2014striving}, can measure the importance of input elements, the Shapley value is proved to be the unique attribution satisfying four desirable properties, \emph{i.e.} the linearity property, the dummy property, the symmetry property, and the efficiency property~\cite{grab1999An}. The four properties can be considered the solid theoretic support for the Shapley value. 
The scientific rigor of the Shapley value makes the attribution analysis and the interaction defined on the Shapley value more trustworthy than other explanation methods.
Please see section~\ref{appendix:a} in supplementary materials for more discussion. 

We have analyzed regional attributions and pixel-wise interactions on different DNNs.
The analysis of regional attributions has demonstrated that important image regions for the $L_2$ attack and those for the $L_\infty$ attack were similar, although $L_2$ perturbations and $L_\infty$ perturbations were significantly dissimilar.
Furthermore, our research has provided new insights into adversarial perturbations  by investigating the property of perturbation components.

\noindent{\textbullet{ Our research has provided a game-theoretic view to explain and verify the phenomenon that adversarailly-trained DNNs mainly focus on foreground. For adversarially-trained DNNs, adversarial perturbations are more likely to interact with each other on the foreground.}}

\noindent{\textbullet{ Moreover, the adversarial-trained DNN usually had more components punishing the true category and less components encouraging incorrect categories than the normally-trained DNN.}}

{In fact, our research group led by Dr. Quanshi Zhang has proposed game-theoretic interactions, including interactions of different orders~\cite{zhang2020game} and multivariate interactions~\cite{zhang2021interpreting}.
As a basic metric, the interaction can be used to learn baseline values of Shapley values~\cite{ren2021learning} and to explain signal processing in trained DNNs from different perspectives. 
For example, we have used interactions to build up a tree to explain the hierarchical interactions between words in NLP models~\cite{zhang2021building} and to explain the generalization power of DNNs~\cite{dropout2020zhang}.
The interaction can also explain adversarial transferability~\cite{wang2021a} and adversarial robustness~\cite{ren2021game}. 
As an extension of the system of game-theoretic interactions, in this study, we interpret attributions and interactions of adversarial attacks.}

However, the computational cost of the Shapley value is NP-hard, which makes the decomposition of perturbation components is also NP-hard. 
Thus, we develop an efficient approximation approach to the decomposition problem.
Our method has been applied to DNNs with various architectures for different tasks.
Preliminary experiments have demonstrated the effectiveness of our method.

\textbf{Contributions:}
This study provides new perspectives to understand adversarial attacks, which includes the regional attribution to the adversarial attack and the extraction of perturbation components.
We have applied our methods to various benchmark DNNs and datasets.
Experimental results provide an insightful understanding of adversarial attacks.

\section{Related work}
\textbf{Adversarial attacks and defense:}
Attacking methods can be roughly divided into two categories, \emph{i.e.} white-box attacks~\cite{intriguing2014,cw2017, advtrain2,bim,pgd2018, jsma,Su2019Onepixel} and black-box attacks~\cite{substitute,transfer_attack,mim,zoo}.
Various methods have been proposed to defend adversarial attacks.
Defensive distillation~\cite{advtrain1} uses knowledge distillation~\cite{knowledge_distillation} to improve the adversarial robustness of DNNs.
Some studies focus on the methods of detecting adversarial examples~\cite{Feinman2017detecting,Metzen2017detection,SafetyNet,attribute_steered_detection}, which can reject adversarial examples in order to protect DNNs.
Besides the detection of adversarial examples, some methods are proposed to directly learn robust DNNs.
Adversarial training methods have been proposed to train DNNs resistant to adversarial attacks~\cite{advtrain3,pgd2018,stability_training}, which use adversarial examples as training samples during the training process.

\textbf{The explanation of adversarial examples:} has received increasing attention in recent years.
Tsipras~\emph{et al.}~\cite{Tsipras2019odds} showed an inherent trade-off between the standard accuracy and adversarial robustness, and found that compared with normally trained DNNs, the adversarial trained DNN tended to be more interpretable.
Furthermore, Etmann~\emph{et al.} theoretically and empirically demonstrated that more robust DNNs exhibited more interpretable saliency maps.
Some studies demonstrated that the existence of adversarial examples was attributed to finite-sample overtting~\cite{Bubeck2018constraints}, the presence of well-generalizing but non-robust features in datasets\cite{not_bugs}, and the geometry property of the high dimensional data~\cite{adv_sphere}.
Wen~\emph{et al.}~\cite{clever} proposed the CLEVER score as the lower bound guarantee of the robustness of DNNs.
Adversarial saliency map~\cite{jsma} computes the importance of each pixel to the network prediction. Xu~\etal~\cite{xu2018structured} and Fan~\etal~\cite{fan2020sparse} proposed to generate structured and sparse perturbations to better understand adversarial examples.
Unlike previous studies of explaining why the adversarial example exists, our work focuses on the explanation of adversarial examples from perspectives of which region of the input image is important to the attacking cost, and how the adversarial perturbation pixels interact with each other during attacking.

\noindent{\textbullet{\textbf{ Difference between the Shapley-based attribution and the adversarial saliency map~\cite{jsma}}:
The gradient-based explanations~\cite{simonyan2013deep}, including the adversarial saliency map~\cite{jsma}, represents the marginal attacking utility \emph{w.r.t.} a specific context.
In comparison, \cite{grab1999An} has proved that the Shapley value is computed on all potential contexts, which ensures more fairness than the gradient-based explanation, and makes the Shapley value satisfy the aforementioned properties.
Please see section~\ref{appendix:a} in supplementary materials for details.}}

\textbf{The interaction} has been widely investigated in the field of statistics~\cite{bien2013lasso,lim2015learning}.
Sorokina~\emph{et  al.}~\cite{Sorokina2018detecting} defined the $K$-$way$ interaction for additive models.
Lundberg~\emph{et al.}~\cite{Lundberg2017tree} quantified interactions between features for tree ensemble methods.
Some studies mainly focus on interactions for DNNs.
Murdoch~\emph{et al.}~\cite{Murdoch2018beyond} proposed to disambiguate interactions in LSTMs, and Singh~\emph{et al.}~\cite{Sigh2019hirachical} extended this method to generic DNNs.
Jin~\textit{et al.}~\cite{jin2020hierachical} quantified the contextual independence of words for DNNs in NLP tasks.
Tsang~\textit{et al.}~\cite{tsang2018detecting} detected statistical interactions based on neural network weights.
Janizek~\textit{et al.}~\cite{janizek2020explaining} extended the method of Integrated Gradients~\cite{sundararajan2017axiomatic} to quantify pairwise interactions of input features.
In comparison, in this study, we apply a different type of interactions between perturbation pixels based on Shapley values, in order to extract perturbation components.

\begin{figure*}[t]
  \centering
  \includegraphics[width=0.9\linewidth]{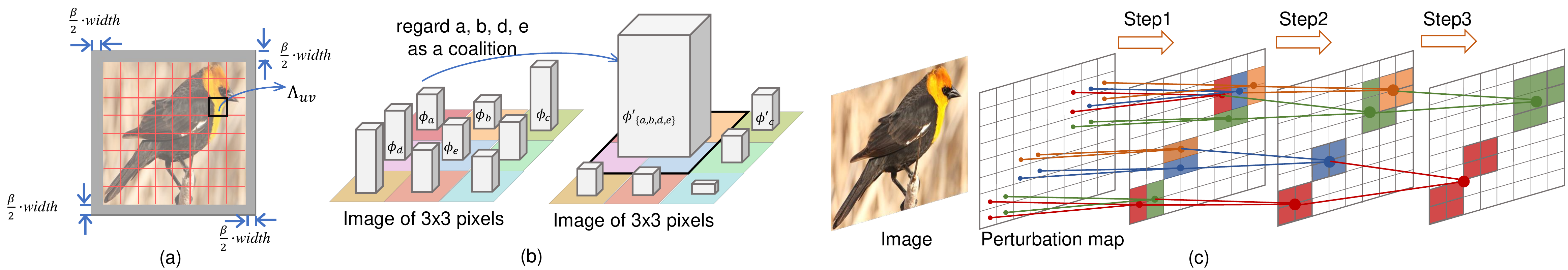}
  \caption{{(a) Images are divided into {\small $L\times L$} grids and extended to {\small $(1+\beta)\cdot width \times (1+\beta)\cdot height$} to compute regional attributions. 
  (b) A toy example to illustrate the interaction among pixels. If pixels $a,b,d,e$ form a coalition and act as a singleton player, then the total reward of $a,b,d,e$ increases. The additional reward indicates the interaction among $a, b,d,e$. 
  (c) A toy example to illustrate the extraction of perturbation components.
}
  } 
  \vspace{-10pt}
  \label{fig:2}
\end{figure*}

\textbf{The Shapley value} is proposed in the game theory~\cite{shapley1953value}. 
Considering multiple players in a game, each player aims to win a high reward.
Some players choose to form a coalition in order to let the coalition win a reward, higher than the sum of rewards of those players when they play individually.
The Shapley value is considered as a unique and unbiased approach to fairly allocating the total reward gained by all players to each player~\cite{lundberg2017unified},
which satisfies four {desirable properties}, \emph{i.e.} \textit{linearity}, \textit{dummy}, \textit{symmetry}, and \textit{efficiency}~\cite{grab1999An},  which will be introduced later. 
Let  $N = \{1, 2, \dots, n\}$ denote the set of all players, and let $r(\cdot)$ denote the reward function.
Let us consider a set of players $S$, which does not include the player $i$, \emph{i.e.}  $S \subseteq N \setminus \{i\}$.
$r(S)$ represents the reward gained by players in $S$, \emph{i.e.} the reward obtained when only players in $S$ participate in the game.
When player $i$ joins the game, the overall reward changes to $r(S\cup\{i\})$.
The difference of the reward, \textit{i.e.} $r(S \cup \{i\}) - r(S)$, is considered as the marginal contribution of player $i$ to the reward.
The Shapley value $\phi^r(i)$ is formulated as the weighted sum of marginal contributions of player $i$ brought by all possible  $S \subseteq N \setminus \{i\}$.
\begin{small}
\begin{equation} \label{eq:shapley}
  \phi^r(i) = \frac{1}{|N|}\sum_{S\subseteq N\setminus \{ i \}}\binom{|N|-1}{|S|}^{-1} (r(S \cup \{i\}) - r(S))
\end{equation}
\end{small}where {\small $\binom{|N|-1}{|S|}$} is the number of all combinations of  $|S|$ players among the set without the player $i$.
Note that the Shapley value is the unique function that satisfies all the  following desirable properties~\cite{grab1999An}:

\textbullet{  Linearity property:} Let there be two games and the corresponding score functions are $r$ and $w$, \emph{i.e.}  $r(S)$ and  $w(S)$ measure the score obtained by players in  $S$ in these two games. If these two games are combined into a new game, and the score function becomes $r+w$, then the Shapley value comes to be $\phi^{r+w}(i)=\phi^r(i) + \phi^w(i)$ for each player.

\textbullet{ Dummy property:} A player $i \in N$ is referred to as a dummy player if  $r(S\cup \{i\}) = r(S) + r(\{i\})$ for any $S \subseteq N \backslash \{i\}$. In this way, $\phi^r(i) = r(\{i\})-r(\emptyset)$, which means that player $i$ plays the game independently. 

\textbullet{ Symmetry property:} If $r(S \cup \{i\}) = r(S \cup \{j\})$ holds for any subset $S\subseteq N\backslash\{i, j\}$, then Shapley values of player $i$ and $j$ are equal, \textit{i.e.} $\phi^r(i)=\phi^r(j)$ .

\textbullet{ Efficiency property:} The sum of each player's Shapley value is equal to the score won by the coalition $N$, \textit{i.e.}  $\sum_{i=1}^n \phi^r(i) = r(N)-r(\emptyset)$. This property guarantees the overall score can be allocated to each player in the game.

\section{Algorithm}
We first introduce basic concepts of adversarial attacks.
Let $x \in [0, 1]^n$ denote the input image, and a DNN is learned to predict the class label $c(x) \in \{1, 2, \dots, T\}$.
To simplify the story, in this study, we mainly analyze the targeted adversarial attack.
The goal is to add a human imperceptible perturbation $\delta \in \mathbb{R}^n$ on $x$ to get a new image $x' = x + \delta$, which makes the DNN mistakenly classify $x'$ as the target category $t\neq c(x)$.
The objective of the targeted adversarial attack is defined by~\cite{cw2017} as follows.
\begin{small}
\begin{equation}
\begin{split}
    &\min_\delta                        \; \| \delta \|_p \quad \text{s.t.} \; c(x+\delta) = t, \;x+\delta \in [0,1]^n \\
  &{\xRightarrow[]{\text{relax}} }    \min_\delta  \; \| \delta \|_p + \lambda \cdot f(x+\delta) \quad \text{s.t.}\; x+\delta \in [0,1]^n   \label{eq:attack}
\end{split}
\end{equation}
\end{small}
where the value of $f(x):[0, 1]^n \rightarrow \mathbb{R}$ measures the correctness of the classification, and $\lambda$ is a scalar constant.
For example, in~\cite{intriguing2014} $f$ is defined as the cross entropy loss.
In~\cite{cw2017}, $f$ is chosen as {\small $f(x) = \max\{\max_{i\ne t} Z_i - Z_t, -\text{threshold}\}$}, where $Z$ is the output of the DNN before the softmax layer.

\subsection{Regional attributions to the adversarial attack}
Given an input image, the regional attribution measures the importance of each image region to the decrease of the attacking cost $\|\delta\|_p$.
\cite{lundberg2017unified} has discussed that the commonly used gradient \emph{w.r.t.} the attacking loss~\cite{simonyan2013deep} cannot objectively reflect the regional attribution due the highly non-linear representation of the DNN. Please see section~\ref{appendix:a} in supplementary materials for more discussions.
Considering the high computational burden, we investigate the regional attribution instead of the pixel-wise attribution.
We divide the entire image into {\small $L \times L$} grids (regions), denoted by {\small $\Lambda = \{\Lambda_{11}, \Lambda_{12}, \dots, \Lambda_{LL}\}$}.
Each region {\small $\Lambda_{uv} \in \Lambda$ ($1 \leq u, v \leq L$)} is a set of pixels.
$\phi_{uv}$ denotes the attribution of the region $\Lambda_{uv}$.
The total attribution to the decrease of the attacking cost is allocated to each region:
\begin{small}
\begin{equation} \label{eq:attribution_sum}
  \phi_{11} + \phi_{12} + \dots + \phi_{LL} =  \text{cost}(\Lambda) - \text{cost}(\emptyset)
\end{equation}
\end{small}where {\small $\text{cost}(\Lambda) = \|\delta^{(\Lambda)}\|_p$} and {\small $\text{cost}(\emptyset) = \|\delta^{(\emptyset)}\|_p$}.
{
{\small $\delta^{(\Lambda)}$} represents the adversarial perturbation generated by allowing all pixels to be perturbed.
{\small $\delta^{(\emptyset)}$} represents the adversarial perturbation generated without perturbing any pixels in $\Lambda$.
Note that it is impossible to conduct adversarial attacks, when all pixels in the image are not allowed to be perturbed.
Thus, as Fig.~{\ref{fig:2}~(a)} shows, we approximate $\phi_{uv}$ by extending the input image to $(1+\beta)\cdot width \times (1+\beta)\cdot height$, where $\beta$ is a small scalar constant.
We regard {\small $\delta^{(\emptyset)}$} as the adversarial perturbation generated when only the extended regions are perturbed.
}
We compute such regional attribution as the Shapley value~\cite{shapley1953value}.
\begin{small}
\begin{equation} \label{eq:attribution}
\phi_{uv} \!=\! \frac{1}{L^2}\!\!\!\!\sum_{S\subseteq \Lambda\setminus \{ \Lambda_{uv} \}}\!\!\!\binom{L^2-1}{|S|}^{-1}\!\!\!\!\!\!\Big[\text{cost}{(S \cup \{\Lambda_{uv}\})} \!-\! \text{cost}{(S)}\Big]
\end{equation}
\end{small}where {\small $S$} denotes a set of regions excluding {\small $\Lambda_{uv}$}.
The {\small $\text{cost}(S)$} is the {\small $L_p$} norm ($p=2$, or $+\infty$) of the adversarial perturbation generated when only regions in {\small $S$} are allowed to be perturbed during attacking.
We formulate such an attack using masking operation, {\small $\hat{\delta} = \arg\min_\delta \; \| \delta \circ M^{(S)} \|_p + c \cdot f(x+\delta\circ M^{(S)}) \ \text{s.t.}\ x+\delta \circ M^{(S)} \in [0,1]^n$}, where $\circ$ represents the element-wise multiplication. {\small $M^{(S)}$} is a mask, which satisfies {\small $ \forall \Lambda_{uv} \in S,\, \forall i \in \Lambda_{uv},\, M^{(S)}_i = 1$}, and for all other pixels $i$, {\small $M^{(S)}_i = 0$}.
{\small $\delta^{(S)} = \hat{\delta}\, \circ\, M^{(S)}$} is the adversarial perturbation generated when only regions in {\small $S$} are allowed to be perturbed, thereby {\small $\text{cost}(S) = \|\delta^{(S)}\|_p$}.

Based on Equation~(\ref{eq:attribution}), in order to quantify the regional attribution, we sample different $M^{(S)}$ to conduct adversarial attacks. 
Note that the adversarial attack with masking operation is conducted to compute $\text{cost}{(S)}$ and $\text{cost}{(S \cup \{\Lambda_{uv}\})}$, instead of obtaining more robust perturbations.

\subsection{Interactions in the attack and the decomposition of perturbation components}
In this section, given a perturbation map of an input image, we aim to decompose the perturbation map into several relatively independent image regions ({perturbation components}), in order to explore the true pixel-wise collaborative behavior behind of the adversarial attack.
Note that the collaborative behavior itself may not be adversarial robust.
Given the set of all perturbation pixels {\small $\Omega$}, we aim to extract $m$ components, \emph{i.e.} {\small $\{C^{(1)}, C^{(2)}, \dots, C^{(m)}\}$}, where {\small $\forall i, C^{(i)}\subseteq \Omega$}; {\small $ \forall i, j, i\ne j, C^{(i)} \cap C^{(j)} = \emptyset$}, {and \small $C^{(1)}\cup C^{(2)} \cup\dots C^{(m)} \subseteq \Omega$}.
Perturbation pixels within each component are supposed to have strong {interactions}, while perturbation pixels in different components are supposed to have weak interactions.
Thus, each component can be roughly considered independent in the adversarial attack.

{
{\textbf{The interaction among perturbation pixels}} reflect cooperative or adversarial relationships of perturbation pixels in attacking.
Inspired by~\cite{grab1999An}, we define interactions based on the {game theory}.
Let us consider a {game} with $n$ {players} {\small $\Omega^\prime=\{1,2,\dots, n\}$}, where players aim to gain a {reward}.
We use {\small $2^{\Omega^\prime}$} to denote all potential subsets of {\small $\Omega^\prime$}. For example, if {\small $\Omega^\prime=\{a, b\}$}, then {\small $2^{\Omega^\prime}=\{\emptyset, \{a\}, \{b\}, \{a, b\}\}$}.
The reward of the game {\small $z: 2^{\Omega^\prime}\rightarrow\mathbb{R}$} maps a set of players to a scalar value.}
Here, given an adversarial example {\small $x^\prime=x+\delta \in \mathbb{R}^n$} and a DNN {\small $g(\cdot)\footnote{$g(\cdot)$ denotes the DNN's output scores before the softmax layer.}:\mathbb{R}^n\rightarrow\mathbb{R}^T$}, where {\small $T$} is the number of categories, we regard each perturbation pixel in $\delta$ as a player.
The goal of perturbation pixels is to decrease the score of the true category $g_{\ell}(x^\prime)$ and increase the score of the target category $g_{t}(x^\prime)$.
Given a set of perturbation pixels {\small $S\subseteq \Omega$}, we formulate the reward of perturbation pixels in {\small $S$} as {\small $z(S) = g_t(x +\delta\circ M^{(S)} ) - g_{\ell}(x + \delta\circ M^{(S)})$}, by assigning values of perturbation pixels in {\small $\Omega\setminus S$} to zero, where {\small $M^{(S)}$} is a mask {\small $\forall i \in S, M^{(S)}_i = 1; \forall i \in \Omega\setminus S, M^{(S)}_i = 0$}.

The total reward in the game is {\small $\Phi = z(\Omega)-z(\emptyset)$, where $z({\Omega}) = g_t(x + \delta \circ {\bf{1}}) - g_{\ell}(x + \delta \circ {\bf{1}})$} is the reward obtained by all perturbation pixels, and {\small $z(\emptyset) = g_t(x + \delta \circ {\bf{0}}) - g_{\ell}(x + \delta \circ {\bf{0}})$} is the baseline reward \emph{w.r.t.} the original image.
The total reward can be allocated to each perturbation pixel, {\small $\Phi = \phi_1 + \phi_2 + \dots + \phi_n$} as the Shapley value.
$\phi_i$ is referred to as the reward of the perturbation pixel $i$.

We notice that perturbation pixels do not contribute to the attack independently.
Instead, some perturbation pixels may form a specific component.
In this way, the reward gained by the component is different from the sum of the reward of each perturbation pixel when they contribute to the attack individually.
Here, we can consider this component as a singleton player.
Thus, the total reward {\small $\Phi$} is allocated to {\small $n-|S|+1$} players in the new game, \emph{i.e.} {\small $\Phi = \phi^\prime_S + \sum_{i\in N\setminus S} \phi^\prime_i$}.
In this way, the interaction among players in {\small $S$} is defined as
{
\begin{equation} \label{eq:interaction}
  I[S] = \phi'_{S} - {\textstyle\sum}_{i\in S}\,\phi_i
\end{equation}
}where {\small $\phi_S^\prime$} is the allocated reward when {\small $S$} is taken as a singleton player, and {\small $\phi_i$} is the reward when we consider the perturbation pixel $i$ contributes to the attack independently.

We compute {\small $\phi_i$} and {\small $\phi'_{S}$} as Shapley values, which will be given in Equation~\eqref{eq:phi_s}.
Fig.~{\ref{fig:2}~(b)} shows a toy example for the interaction.
The total reward is allocated to each perturbation pixel, \emph{i.e.} {\small $\Phi = \phi_a + \phi_b + \dots + \phi_h + \phi_i$}.
If pixels $a,b,d,e$ are regarded as a singleton player, then the allocation of {\small $\Phi$} would change to {\small $\Phi = \phi'_{\{a, b,d,e\}} + \phi'_c \dots + \phi'_h + \phi'_i$}.
If {\small $\phi'_{\{a, b,d,e\}} - (\phi_a + \phi_b + \phi_d + \phi_e) \ne 0$}, then pixels $a, b, d,e$ are considered to have interactions.

\textit{Understanding of the interaction:}
If {\small $I[S] > 0$}, it means perturbation pixels in {\small $S$} cooperate with each other to change prediction scores of the DNN.
If {\small $I[S] < 0$}, it means perturbation pixels in {\small $S$} conflict with each other.
The absolute value {\small $|I[S]|$} indicates the strength of the interaction.

\textit{Computation of the reward:}
According to Equation~\eqref{eq:shapley}, {\small $\phi_i$} and {\small $\phi'_{S}$} are computed as follows.
{
\small 
\begin{equation}
\begin{split}
    \phi_{i}   &= \frac{1}{n}\sum_{\tilde{S}\subseteq \Omega\setminus \{ i \}}\binom{n-1}{|\tilde{S}|}^{-1}\Big[z(\tilde{S}\cup \{i\}) - z(\tilde{S})\Big] \\
  \phi'_{S}  &= \frac{1}{n^\prime}\sum_{\tilde{S}\subseteq \Omega\setminus S}\binom{n^\prime-1}{|\tilde{S}|}^{-1}\Big[z(\tilde{S}\cup S) - z(\tilde{S})\Big] \label{eq:phi_s}
\end{split}
\end{equation}
}where {\small $n^\prime = n-\left|S\right|+1$}.
In this study, we propose an efficient method to approximate {\small $\phi_i$} and {\small $\phi'_{S}$}, which will be introduced in Equation~\eqref{eq:taylor_a} and Equation~\eqref{eq:taylor_b}.

\textbf{Extraction of perturbation components:}
We extract perturbation components via hierarchical clustering, in which perturbation pixels have strong interactions.
The pseudo code of extracting perturbation components is shown in section~\ref{appendix:c} in supplementary materials.
In the first step of clustering, we merge $q$ neighboring perturbation pixels with strong interactions into a $q$-pixel component.
In the second step, we merge $q$ neighboring components with strong interactions into a component of $q^2$ pixels.
In this way, we iteratively generate components of $q$, $q^2$, $q^3$ pixels, \textit{etc.}
A toy example is shown in Fig.~\ref{fig:2}~(c).
We use {\small $C^{(i)}$} to denote a component.
The finally merged component is selected from a set of component candidates, each of which is a set of $q$ neighboring components, {\small $S^{c} = C^{(i_1)}\cup C^{(i_2)}\cup\dots\cup C^{(i_q)}$}.
If {\small $\left |I[S^c]\right | >\gamma$}, then we merge {\small $C^{(i_1)}, C^{(i_2)}, \dots, C^{(i_q)}$} into a large component.
Considering the local property~\cite{Chen2018L}, we only compute interactions among neighboring pixels/components.

\textbf{Efficient approximation of rewards of perturbation pixels:}
We can consider the value of each perturbation pixel $i$ as the sum of values of $K$ sub-pixels, denoted by $(i, k)$,  $1 \le k \le K$.
The values of the $K$ sub-pixels are uniformly divided, \emph{i.e.} {\small $\delta/K = \delta_{(i, 1)} = \delta_{(i, 2)} = \dots = \delta_{(i, K)}$}.
In this way, the reward $\phi_i$ can be approximated as the sum of sub-pixels' rewards, \textit{i.e.} $\phi_i \approx \sum_{k=1}^{K}\phi_{(i, k)}$.
Consider the symmetry property of the Shapley value~\cite{grab1999An}, $\phi_{(i, 1)} = \phi_{(i, 2)} = \dots = \phi_{(i, K)}$.
Thus, we can approximate $\phi_{i} \approx K\cdot \phi_{(i, k)}$.
Please see section~\ref{appendix:b.1} in supplementary material for the proof and more discussions.
The reward $\phi_{(i,k)}$ can be approximated as follows.
Let us consider the marginal reward for computing the Shapley value of the sub-pixel, \textit{i.e.} {\small $z{(S \cup \{(i, k)\})} - z{(S)}$}.
Given a large value of $K$, the perturbation magnitude of each sub-pixel $|\delta_{(i, k)}|$ is fairly small.
The marginal reward can be approximated as {\small $z{(S \cup \{(i, k)\})} - z{(S)} = \delta_{(i, k)}\cdot \partial{z(S)}/ \partial{\delta_{(i, k)}} + {o(\delta_{(i, k)})}$}, based on the Taylor expansion.
In this way, the Shapley value of each sub-pixel is given as Equation~\eqref{eq:taylor_a}.
\begin{small}
\begin{equation}\label{eq:taylor_a}
    \phi_{(i, k)} \!\approx\! \frac{1}{nK}\!\!\!\sum_{S\subseteq \Omega^{\text{pixel}}\setminus \{ (i, k) \}}\!\!\!\!\binom{nK-1}{|S|}^{\!\!-1} \!\!\!\!\!\!\!\!\!\!\!\!\!\!\!\underbrace{\left[\frac{\partial z(S)}{\partial \delta_{(i, k)}}\delta_{(i, k)}\right]}_{{\textrm{approximates}\; z(S\cup \{(i, k)\}) - z(S)}}\!\!\!\!\!\!\!\!\!\!\!\!\!\!\!
\end{equation}where {\small $\Omega^{\text{pixel}} = \{(i, k)|1 \leq i \leq n, 1\leq k\leq K\}$}.
\end{small}

\textbf{Efficient approximation of rewards of components:}
Let there be $m$ components in a certain clustering step, \emph{i.e.} {\small $\{C^{(1)}, C^{(2)}, \dots, C^{(m)}\}$}.
Given a component {\small $C^{(u)}$}, 
we approximate {\small $\phi_{C^{(u)}}$} using combinations of components, instead of perturbation pixels.
We further reduce the cost of calculating {\small $\phi_{C^{(u)}}$} in the similar way of calculating $\phi_i$.
Each perturbation pixel $i$ in {\small $C^{(u)}$} is divided into $K$ sub-pixels, \textit{i.e.} {\small $\forall i\in C^{(u)}, \delta_i = \sum_k \delta_{(i, k)}$}.
In this way, $C^{(u)}$ can be divided into $K$ sub-components.
The $k$-th sub-component is given as {\small ${C^{(u)}_k} = \bigcup_{i \in C^{(u)}} \{(i, k)\}$}.
The reward of {\small $C^{(u)}_k$} is approximated as follows.
Please see section~\ref{appendix:b.3} in supplementary materials for the proof.
{
\small
\begin{equation}
\begin{split}
\phi_{C^{(u)}_k} \approx \frac{1}{m K}\sum\nolimits_{S\subseteq \Omega^{\text{comp}}\setminus \{C^{(u)}_k\}} \binom{m K - 1}{|S|}^{-1}  A \label{eq:taylor_b}
\end{split}
\end{equation}
}where {\small $\Omega^{\text{comp}} = \{C^{(u)}_k|1\le u\le m, 1\leq k\leq K\}$}, and {\small $A=\sum_{(i, k)\in C^{(u)}_k}\left[\frac{\partial z(S)}{\partial \delta_{(i, k)}}\delta_{(i, k)}\right]$}, which approximates {\small $z(S\cup\{C^{(u)}\})-z(S)$}.

\textbf{Implementation \& computational complexity:}
The computation of Shapley values is NP-hard.
The sampling-based method has been widely used for approximation~\cite{castro2009polynomial}.
Thus, we propose to approximate Equation~\eqref{eq:taylor_a} and Equation~\eqref{eq:taylor_b} by a sampling method.
The complexity of computing the Shapley value of one {pixel} is $O(2^n)$.
Equation~\eqref{eq:sample_pixel} reduces the computational complexity to $O(nKT)$, where $n$ is the pixel number, and $T$ is times of sampling.
Because derivatives to sub-pixels \emph{w.r.t.} all pixels {\small $i \in \Omega$} can be computed simultaneously, the computational complexity of computing Shapley values of all {pixels} remains $O(nKT)$.
{
  \small
  \begin{equation}
      \begin{split}
           \phi_{(i, k)}\! \approx& \frac{1}{nKT}\!\! \sum_{s=0}^{nK-1}\sum_{t=1}^{T} \delta_{(i, k)} \left.\frac{\partial z(S)}{\partial \delta_{(i, k)}}\right\vert_{S=S_{st}^{\text{pixel}}} \!\!s.t.  \left\vert S_{st}^{\text{pixel}}\right\vert=s \!\!\!\!\!\!\!\! \!\!\!\!\label{eq:sample_pixel} \\
           \phi_{C^{(u)}_k}\! \approx& \frac{1}{mKT} \sum_{s=0}^{mK-1}\!\!\sum_{(i, k)\in C^{(u)}_k}\sum_{t=1}^{T} \delta_{(i, k)} \left.\frac{\partial z(S)}{\partial \delta_{(i, k)}}\right\vert_{S=S_{st}^{\text{comp}}}\!\!\!\!\!\!\!\!\!\!\!\! \\
           &s.t. \quad \left\vert S_{st}^{\text{comp}}\right\vert =s
      \end{split}
  \end{equation}
}where $t$ represents a sampling step; $s$ controls the size of the set $S$; {\small $S_{st}^{\text{pixel}} \subseteq \Omega^{\text{pixel}} \setminus \{(i, k)\}$} denotes a random subset of $s$ sub-pixels excluding $(i, k)$; {\small $S_{st}^{\text{comp}} \subseteq \Omega^{{\text{comp}}} \setminus \{C^{(u)}_k\}$} denotes a random subset of $s$ sub-components excluding  {\small $C^{(u)}_k$}.

Even with above approximation, the computational cost for computing all component candidates in each step of clustering is still high.
Thus, we further approximate rewards of component candidates by simplifying contextual relationships of far-away pixels~\cite{Chen2018L}.
We use {\small $S^{c} = C^{(i_1)}\cup C^{(i_2)}\cup\dots\cup C^{(i_q)}$} to denote a component candidate.
If we compute $\phi^\prime_{S^c}$ in the set {\small \{$S^c, C^{(i_{k+1})} \dots, C^{(i_m)}\}$}, the computational complexity is $O((m-q)KT)$.
The complexity of computing rewards of all candidates is $O(m(m-q)KT)$.
Here, instead of computing {\small  $\phi^\prime_{S^c}$} in the set {\small \{$S^c, C^{(i_{k+1})} \dots, C^{(i_m)}\}$}, we randomly merge $\tilde{m}$ components to get $\tilde{m}/q$ component candidates, including $S^c$.
In this way, the new set includes $\tilde{m}/q$ component candidates and $m-\tilde{m}$ components, \emph{i.e.} {\small $\{S^c, \bigcup_{a=q+1}^{2q}C^{(i_a)}, \dots, \bigcup_{a=\tilde{m}-q+1}^{\tilde{m}}C^{(i_a)}, C^{(i_{\tilde{m}+1})}, \dots, C^{(i_m)}\}$.}
We can simultaneously compute rewards of $\tilde{m}/q$ candidates in the new set, and the computational complexity is {$O((m-(q-1)\tilde{m}/q)KT)$}.
To compute rewards of all potential component candidates, we need to sample $qm / \tilde{m}$ different sets.
In this way, the overall complexity for the computation of rewards of candidates is reduced from {$O(m(m-q)KT)$} to {$O(m(qm/\tilde{m}-q)KT)$}.
Please see section~\ref{appendix:b.4} supplementary materials for details.

\begin{figure*}[t]
  \centering
  \includegraphics[width=0.88\linewidth]{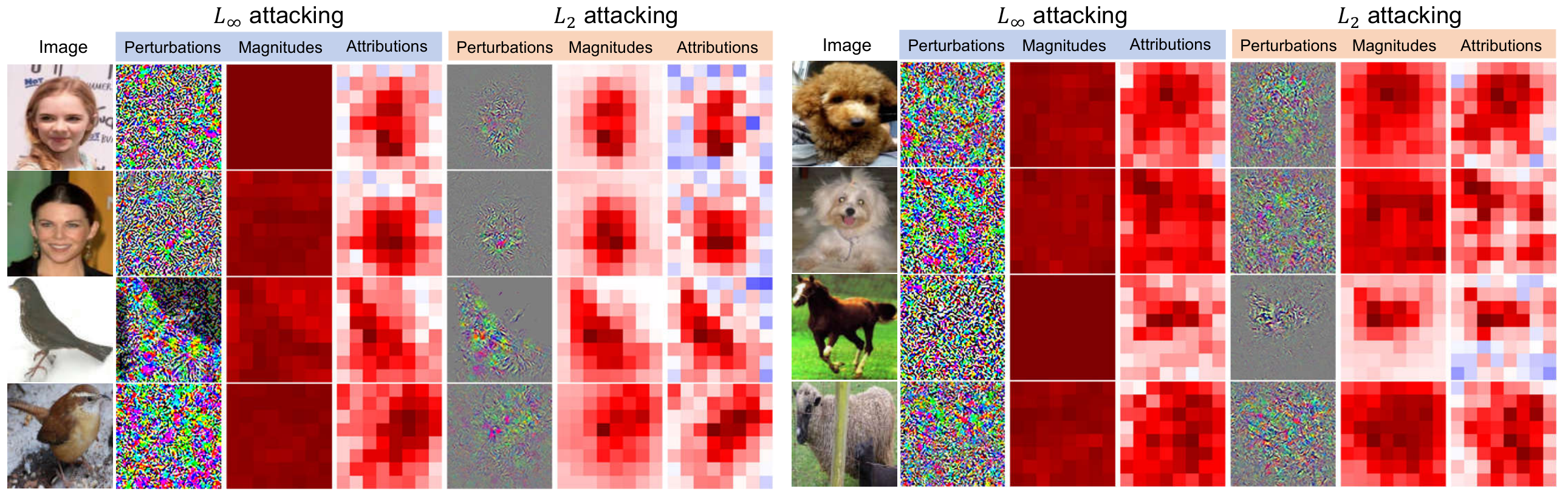} 
  \vspace{-3pt}
  \caption{Comparison between adversarial perturbations and regional attributions. For attributions, dark red regions are important regions to the decrease of the attacking cost. 
  Blue regions indicate negative attributions, \emph{i.e.} these regions decrease the effect of attacking and increase the attacking cost.
  Regional attributions to the $L_2$ attack and magnitudes of regional perturbations were usually similar, while regional attributions to the $L_\infty$ attack and magnitudes of regional perturbations were usually different.
    }
  \label{fig:exp_attribution} 
    \vspace{-5pt}
\end{figure*}

\begin{table*}[t]
  \centering
  \resizebox{0.88\linewidth}{!}{
    \begin{tabular}{|c|c|c|c|c|c|c|c|c|}
      \hline
                & \multicolumn{2}{|c|}{CelebA} & \multicolumn{2}{|c|}{CUB200-2011} & \multicolumn{2}{|c|}{Pascal VOC} & \multicolumn{2}{|c|}{Stanford Dogs}                                                                                   \\ \cline{2-9}
                & IoU(attributions)              & IoU(magnitudes)                          & IoU(attributions)                  & IoU(magnitudes)                            & IoU(attributions)       & IoU(magnitudes)        & IoU(attributions)         & IoU(magnitudes)        \\ \hline
      AlexNet   & \textbf{0.749 $\pm$ 0.027} & {0.668 $\pm$ 0.154} & \textbf{0.729 $\pm$ 0.032} & {0.526 $\pm$ 0.051} & \textbf{0.743 $\pm$ 0.046} & {0.577 $\pm$ 0.069} & \textbf{0.739 $\pm$ 0.024} & {0.567 $\pm$ 0.061} \\ \hline
      VGG-16    & \textbf{0.718 $\pm$ 0.038} & {0.534 $\pm$ 0.064} & \textbf{0.681 $\pm$ 0.057} & {0.543 $\pm$ 0.070} & \textbf{0.685 $\pm$ 0.073} & {0.474 $\pm$ 0.096} & \textbf{0.742 $\pm$ 0.045} & 0.488 $\pm$ 0.138 \\ \hline
      ResNet-18 & \textbf{0.762 $\pm$ 0.087} & 0.497 $\pm$ 0.190 & \textbf{0.713 $\pm$ 0.044} & 0.552 $\pm$ 0.064 & \textbf{0.718 $\pm$ 0.072} & 0.540 $\pm$ 0.158 & \textbf{0.707 $\pm$ 0.060} & 0.550 $\pm$ 0.110 \\ \hline
    \end{tabular}
  }
  \caption{
  The IoU between regional attributions to the $L_2$ attack and regional attributions to the $L_\infty$ attack, and the IoU between $L_2$ adversarial perturbations' magnitudes and $L_\infty$ adversarial perturbations' magnitudes\protect\footnotemark[2]. Regional attributions to the $L_2$ attack and regional attributions to the $L_\infty$ attack were similar, while regional magnitudes of $L_2$ adversarial perturbations and magnitudes of $L_\infty$ adversarial perturbations were dissimilar.}

  \label{tab:exp_attribution} 
  \vspace{-10pt}
\end{table*}

\begin{figure*}[t]
  \centering
  \includegraphics[width=0.8\linewidth]{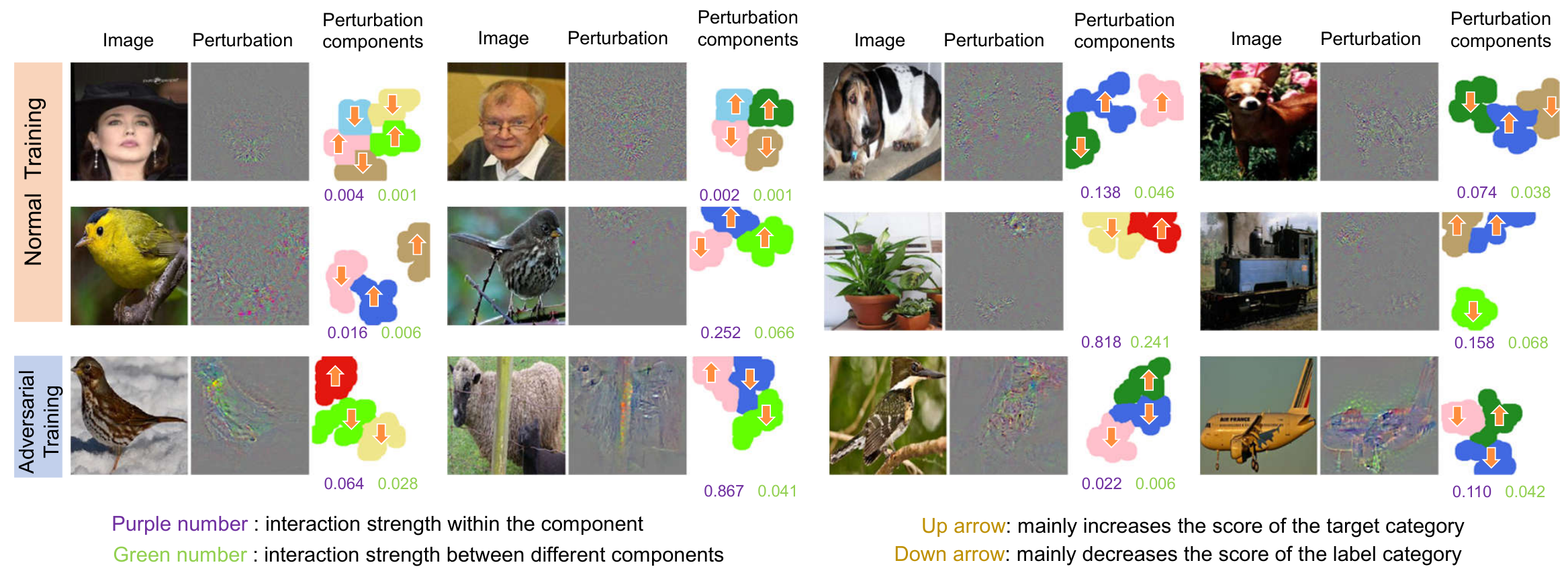} 
  \caption{{Visualization of perturbation components.
      Perturbation pixels within each component have strong interactions. 
      Perturbation pixels between different components have relatively weak interactions.
      The color saturation of the component indicates the average reward (importance) of the perturbation component.
      Perturbation components were not always aligned with visual concepts.
    }} 
    \vspace{-10pt}
  \label{fig:exp_interaction}
\end{figure*}

\section{Experiments}   
\textbf{Datasets \& DNNs:}
We tested our methods on tasks of coarse-grained image classification, fine-grained image classification, and face attribute estimation, using four benchmark datasets: the Pascal VOC 2012 dataset~\cite{voc}, the CUB200-2011 dataset~\cite{cub}, the Stanford Dog dataset~\cite{sdd}, and the CelebA dataset~\cite{celeba}.
For each of these datasets, we used object images cropped by object bounding boxes for both training and testing.
We analyzed regional attributions to the adversarial attack on three benchmark DNNs: AlexNet~\cite{krizhevsky2012imagenet}, VGG-16~\cite{simonyan2015very} and ResNet-18~\cite{he2016deep}.
We computed interactions among perturbation pixels and extracted perturbation components on ResNet-18/34/50~\cite{he2016deep}.

Furthermore, to analyze the utility of adversarial training, we also extracted perturbation components on adversarially-trained ResNet-18/34/50~\cite{pgd2018}.
We conducted adversarial training on two datasets: the Pascal VOC 2012 dataset~\cite{voc} and the CUB200-2011 dataset~\cite{cub}.
For adversarial training on the CUB200-2011 dataset, we only conducted the classification on 10 categories that were uniformly selected from the original 200 categories.
For fair comparisons, the normally-trained DNN was also trained using the same 10 categories in the CUB200-2011 dataset.
All DNNs were pre-trained on the ImageNet dataset~\cite{imagenet2015}, and then fine-tuned using these four datasets, respectively.

\textbf{Implementation details:}
We used the C\&W  attack~\cite{cw2017} as the $L_2$ attacker and the BIM~\cite{bim} as the $L_\infty$ attacker, which have been widely used.
For adversarial attacks on all datasets and DNNs, we conducted the targeted attack.
For the Pascal VOC 2012 dataset~\cite{voc}, we set the target class as the \textit{bird}.
For the CUB200-2011 dataset~\cite{cub} and the Stanford Dogs dataset~\cite{sdd}, we set the target class as the last category in each dataset, which were \textit{the Common Yellowthroat} and \textit{the African hunting dog}, respectively.
For the CelebA dataset~\cite{celeba}, we chose two global face attributes (\textit{male, young}) and three local face attributes (\textit{wearing glasses, smiling, wearing lipstick}), and set the target as the opposite attribute, for example, we conducted the adversarial attack on the CelebA dataset to make a DNN mistakenly predict a face image actually with \textit{wearing glasses} to  \textit{not wearing glasses}.
We set $\beta=1/6$, $L=8$ to compute regional attributions.
We set $q=4$ to compute interactions among perturbation pixels and further extract perturbation components.

\subsection{Exp. 1: Regional attributions to the adversarial attack}
\textbf{Visualization of regional attributions:}
Fig.~\ref{fig:exp_attribution} visualizes adversarial perturbations and regional attributions.
We compared magnitudes of regional adversarial perturbations with regional attributions.
Given a region $\Lambda_{uv}$, the magnitude of the regional perturbation was computed as {\small $(\sum_{i \in \Lambda_{uv}}|\delta_i|^2)^{1/2}$}.
Fig.~\ref{fig:exp_attribution} shows that attributions to the $L_2$ attack and magnitudes of regional perturbations were similar, but attributions to the $L_\infty$ attack and magnitudes of regional perturbations were usually different. 
For example, let us focus on the top left girl image in Fig.~\ref{fig:exp_attribution}, $L_\infty$ perturbations uniformly distributed over the entire image, while attributions mainly focused on the face region. We have also compared regional attributions with different hyper-parameters ($\beta$ and $L$), which shows that regional attributions were insensitive to the selection of hyper-parameters. Please see section~\ref{appendix:d.1} supplementary materials for details.

\textbf{Similarity between regional attributions through different attacks:}
As Fig.~\ref{fig:exp_attribution} shows, although the distribution of $L_2$ adversarial perturbations and the distribution of $L_\infty$ adversarial perturbations were dissimilar, their regional attributions were similar to each other.
Given regional attributions to the $L_2$ attack {\small $\phi^{(2)}\in\mathbb{R}^{L\times L}$} and regional attributions to the $L_\infty$ attack {\small $\phi^{(\infty)}\in\mathbb{R}^{L\times L}$}, we used the {\small $IoU=\frac{\sum_{u}\sum_{v}\min(\phi^{\prime(2)}_{u v}, \phi^{(\prime\infty)}_{u v})}{\sum_u \sum_v\max(\phi^{\prime(2)}_{u v}, \phi^{\prime(\infty)}_{u v})}$} to measure the similarity between regional attributions $\phi^{(2)}$ and
$\phi^{(\infty)}$, where we normalized the attribution {\small $\phi'_{u v} = \frac{\phi_{u v}-\min_{u', v'}\phi_{u' v'}}{\max_{u', v'}\phi_{u' v'} - \min_{u', v'}\phi_{u' v'}}$}.
We also showed the IoU between $L_2$ magnitudes and $L_\infty$ magnitudes.
Please see Table~\ref{tab:exp_attribution} for quantitative results of similarity between regional attributions and similarity between magnitudes of regional perturbations.
In Fig.~\ref{fig:exp_attribution}, we found that regional attributions to the $L_2$ attack and regional attributions to the $L_\infty$ attack were similar, {while magnitudes of $L_2$ adversarial perturbations and regional magnitudes of $L_\infty$ adversarial perturbations were dissimilar.}

\footnotetext[2]{{The mean and standard deviation of IoU were computed on 10 samples.}}

\subsection{Exp. 2: Interactions in the attack and the decomposition of perturbation components}
\textbf{Extraction of perturbation components:}
We extracted interactions between perturbation pixels generated in the $L_2$ attack, and decomposed the perturbation into components.
To reduce the computation cost, we regarded each group of neighboring $4\times 4$ pixels as a super-pixel.
We set $\tilde{m}=0.5\cdot m$.
In the first step of clustering, $\gamma$ was set to satisfy {\small $\mathbb{E}_{S}[{\mathbbm{1}}(\left\vert I[S]\right\vert > \gamma)]=0.2$}, where ${\mathbbm{1}(\cdot)}$ is the indicator function.
In subsequent steps of clustering, $\gamma$ was set to satisfy {\small $\mathbb{E}_{S}[{\mathbbm{1}}(\left\vert I[S]\right\vert > \gamma)]=0.5$}.
{To obtain stable results, when computing the reward of a component candidate $S_c$, we kept the nearest component of each component in $S_c$ always present in sampling.}
In each step, we ended up merging components, when component candidates for which the reward had been computed covered more than 90\% components.
We used the greedy strategy to merge components, and kept conducting clustering until the size of each component was 64.

We first extracted perturbation components on normally-trained DNNs.
Fig.~\ref{fig:exp_interaction} visualizes adversarial perturbations and corresponding perturbation components.
{Note that we enlarged the size of each super-pixel, to clarify the visualization.}
Fig.~\ref{fig:exp_interaction} shows that components were not aligned with visual concepts.
For example, adversarial perturbations on the top left girl image in the first row of Fig.~\ref{fig:exp_interaction} was segmented into five components, in which the pink component covered the neck, hair, and other regions without semantic meanings.

\textbf{Effects of adversarial training on perturbation components:}
We also extracted perturbation components on adversarially-trained DNNs, which are visualized in Fig.~\ref{fig:exp_interaction}.
We used the method proposed by Madry \emph{et al.}~\cite{pgd2018} for adversarial training, which was formulated as {\small $\min _{\boldsymbol{\theta}} \mathbb{E}_{x_i\in X} \max _{x_i':\left\|\mathbf{x}'_{i}-\mathbf{x}_{i}\right\|_p\leq \epsilon} \ell\left(g_{\boldsymbol{\theta}}\left(\mathbf{x}'_{i}\right), c_i\right) $}, where $c_i$ represents the label of $x_i$, and $\ell$ denotes the classification loss.

{
We investigated whether perturbation components were mainly localized in the foreground or the background.
}
A perturbation component was regarded being in the foreground, if perturbation pixels in the component belonging to the foreground were more than perturbation pixels in the component belonging to the background.
Table~\ref{tab:foreground} reports the ratio of components in the foreground for adversarially-trained DNNs and normally-trained DNNs. 
Adversarially-trained DNNs had more perturbation components in the foreground than normally-trained DNNs.

{
Moreover, we also explored the utility of perturbation components, \emph{i.e.} whether the component mainly decreased the prediction score of the true category or mainly increased the score of the target category.
}
We classified perturbation components into two types, \emph{i.e.} components mainly decreasing the prediction score of the true category and components mainly increasing the score of the target category. 
{\small $\Delta y_\ell = \left| g_\ell(x+\delta) - g_\ell(x+\delta \circ M^{(\Omega\setminus C^{(u)})}) \right|$} and {\small $\Delta y_t = \left|g_t(x+\delta) - g_t(x+\delta \circ M^{(\Omega\setminus C^{(u)})}) \right|$} measured the decrease of the prediction score of the true category and the increase of  the prediction score of the target category caused by the component {\small $C^{(u)}$}, respectively.
If {\small $\Delta y_\ell > \Delta y_t$}, the component {\small $C^{(u)}$} was regarded mainly decreasing the score of the true category;
otherwise, mainly increasing the score of the target category.
Table~\ref{tab:effect} reports the ratio of components that mainly decreased the prediction score of the true category.
{The ratio of components mainly decreasing the score of the true category in adversarially-trained DNNs were usually greater than that in normally-trained DNNs.
}

\begin{table}[t]
  \centering
  \resizebox{0.9\linewidth}{!}{
    \begin{tabular}{|c|c|c|c|c|}
      \hline
                 & \multicolumn{2}{|c|}{CUB200-2011} & \multicolumn{2}{|c|}{Pascal VOC}                                                                                  \\ \cline{2-5}
                & Normal & Adv-trained & Normal & Adv-trained        \\ \hline
      ResNet-18   & {63.8\%} & \textbf{80.8\%} & {55.6\%} & \textbf{90.4\%} \\ \hline
      Resnet-34  & 69.0\% & \textbf{78.6\%} & 66.2\% & \textbf{85.7\%}\\ \hline
      ResNet-50  & 62.6\% & \textbf{84.9\%} & 60.8\% & \textbf{89.2\%} \\ \hline
    \end{tabular}
  }
  \vspace{1pt}
  \caption{The ratio of perturbation components mainly in the foreground. 
  }
  \label{tab:foreground} 
  \vspace{-10pt}
\end{table}

\begin{table}[t]
  \centering
  \resizebox{0.9\linewidth}{!}{
    \begin{tabular}{|c|c|c|c|c|}
      \hline
                 & \multicolumn{2}{|c|}{CUB200-2011} & \multicolumn{2}{|c|}{Pascal VOC}                                                                                  \\ \cline{2-5}
                & Normal & Adv-trained & Normal & Adv-trained        \\ \hline
      ResNet-18   & {62.8\%} & \textbf{77.9\%} & {34.9\%} &\textbf{55.3\%} \\ \hline
      ResNet-34    & {44.0\%} & \textbf{68.4\%} & {43.8\%} & \textbf{63.1\%}\\ \hline
      ResNet-50 & 58.2\% &\textbf{82.8\%} & 31.6\%  &  \textbf{48.2\%}\\ \hline
    \end{tabular}
  }
  \vspace{1pt}
  \caption{The ratio of perturbation components which mainly decrease the score of the true category. 
  }
  \label{tab:effect} 
  \vspace{-10pt}
\end{table}

\section{Conclusion}
In this paper, we have analyzed adversarial attacks from the attributional perspective.
We have computed regional attributions to adversarial attacks.
We have further defined and extract interactions among perturbation pixels decomposed the perturbation map into perturbation components based on interactions.
We have found regional attributions to the $L_2$ attack and magnitudes of the $L_2$ perturbation were similar, while regional attributions to the $L_\infty$ attack and magnitudes of the $L_\infty$ perturbation were dissimilar.
The extraction of perturbation components showed that perturbation components were not aligned with visual concepts. 
We have found that adversarially-trained DNNs had more perturbation components in the foreground than normally-trained DNNs.
Moreover, compared to the normally-trained DNN, the adversarially-trained DNN was prone to decrease the score of the true category, instead of increasing the score of the target category.
Our methods have been used to analyze different DNNs learned for the image classification and the face attribute estimation.

\section*{Acknowledgments}
{This work is partially supported by the National Nature Science Foundation of China (No. 61906120, U19B2043), Shanghai Natural Science Foundation (21ZR1434600), Shanghai Municipal Science and Technology Major Project (2021SHZDZX0102), and Huawei Technologies Inc. Xin Wang is supported by Wu Wen Jun Honorary Doctoral Scholarship, AI Institute, Shanghai Jiao Tong University.}

{\small
\bibliographystyle{ieee_fullname}
\bibliography{egbib}
}

\clearpage
\onecolumn
\appendix

\section{Comparisons of Shapley-based attributions and other explanation methods}
\label{appendix:a}
{We define regional attributions and interactions between perturbation pixels based on Shapley values~\cite{shapley1953value}. We compare Shapley-based attributions with other explanation methods from the following perspectives.}

{
\textbullet{ Theoretical rigor.} A good attribution method must satisfy certain desirable properties. The Shapley value has been proved to be the unique attribution that satisfies four desirable properties, \emph{i.e.} the linearity property, the dummy property, the symmetry property, and the efficiency property~\cite{grab1999An}.  In comparison, some explanation methods like Grad-CAM~\cite{selvaraju2017grad} and GBP~\cite{springenberg2014striving} do not have theoretic supports for the correctness of these methods.}

{
\textbullet{ Objectivity.}  The attribution of one input element depends on contexts of neighboring pixels. The Shapley value considers all possible contexts to compute the attribution of an input unit, which ensures the objectiveness of the attribution. In contrast, some attention methods, such as the adversarial saliency map~\cite{jsma}, only consider the marginal gradient, which is biased to a specific context from this perspective.}

{
\textbullet{ Trustworthiness.} The theoretic foundation in game theory makes the Shapley values trustworthy. In contrast, some seemingly transparent explanation methods simply do not have clear theoretical support, which hurts the trustworthiness of the explanation. Actually, \cite{adebayo2018sanity} has shown some explanation methods like GBP~\cite{springenberg2014striving} can not reflect the true attribution.}

{
\textbullet{ Broad applicability.} The Shapley values can be extended to measure interactions between two input elements. \cite{zhang2021building} has proved the theoretical foundation and advantages for defining interactions using the Shapley value in game theory~\cite{grab1999An}. However, some gradient-based explanation methods assume the model is locally linear, which fails to measure interactions between two input elements.}

\section{Details of efficient approximation of interactions}
We approximate Shapley values to enable efficient computation of interactions.
\subsection{Approximation of attributions of perturbation pixels}
\label{appendix:b.1}

In Section Algorithm, we introduce the approximation of the Shapley value of a perturbation pixel.
In the supplementary material, we give more discussions about the approximation.
 
The adversarial perturbation is denoted as $\delta \in \mathbb{R}^n$.
Each perturbation pixel $i$ is divided into $K$ sub-pixels with equal values, \emph{i.e.} $\delta_i = \delta_{(i, 1)} + \delta_{(i, 2)} + \dots + \delta_{(i, K)}$ and $\delta_{(i, 1)} = \delta_{(i, 2)} = \dots = \delta_{(i, K)}$.
Instead of directly computing the attribution of each perturbation pixel, we compute the attribution of each sub-pixel. 
The attribution of the sub-pixel can be efficiently approximated based on the Taylor expansion, which will be discussed later.

Among sub-pixels $(i, 1), (i, 2) \dots (i, K)$, each sub-pixel plays the same role in attacking, thereby $\phi_{(i,1)}= \phi_{(i, 2)} = \dots = \phi_{(i, K)}$, which is proved as follows. 
The attribution of each sub-pixel $(i, k)$ is formulated as the Shapley value.
The Shapley value satisfies the four axioms (linearity axiom, dummy axiom, symmetry axiom, and efficiency axiom).
According to the symmetry axiom, given two sub-pixels $(i, k)$ and $(j, k')$ , if $z(S\cup \{(i, k)\}) = z(S\cup \{(j, k')\})$ holds for any set $S\subseteq\Omega^{\text{pixel}}\backslash \{(i, k), (j, k')\}$, then $\phi_{(i, k)} = \phi_{(j, k')}$, where $\Omega^{\text{pixel}} = \{(1, 1), (1, 2), \dots$, $ (n, K-1), (n, K)\}$ denotes the set of all sub-pixels.
Because the sub-pixel of the same perturbation pixel $i$ has the equal value, given two sub-pixels $(i, k)$ and $(i, k')$ of the same perturbation pixel $i$, $z(S\cup \{(i, k)\}) = z(S\cup \{(i, k')\})$ holds for any set $S\subseteq\Omega^{\text{pixel}}\backslash \{(i, k), (i, k')\}$, where $1 \leq k, k' \leq K$, and $k\neq k'$. 
In this way, $\phi_{(i, 1)} = \phi_{(i, 2)} = \dots = \phi_{(i, K)}$. 
Thus, we approximate the attribution of perturbation pixel $i$ as $\phi_i = \sum_{i=1}^K\phi_{(i, k)}$, which equals to $\phi_i = K\cdot \phi_{(i, k)}$.

\subsection{Properties of the approximated attribution}
\label{appendix:b.2}
In Section Algorithm, we approximate the attribution of perturbation pixel $i$ as $\phi_i = \sum_{i=1}^K\phi_{(i, k)}$. In the supplementary material, we further discuss properties of the approximated attribution.

The approximated attribution still satisfies the linearity axiom and the efficiency axiom. 

\textit{Proof of the linearity axiom:} Given two score functions $v(S)$ and $w(S)$, we use $\phi_i^v$ and $\phi_i^w$ to denote the attribution of perturbation pixel $i$ to score $v$ and score $w$ respectively.
  Let there be a new score function $f'(S) = v(S)+ w(S)$. 
We use $\phi^{v+w}_i$ to denote the approximated attribution of perturbation pixel $i$ to the new score function.
The approximated attribution of perturbation pixel $i$ is the sum of attributions sub-pixels, \emph{i.e.} $\phi^{v+w}_i = \sum_{k=1}^K \phi^{v+w}_{(i, k)}$. 
The attribution of each sub-pixel is defined as the Shapley value. 
The Shapley value satisfies the linearity axiom. 
Then  $ \sum_{k-1}^K \phi^{v+w}_{(i, k)} = \sum_{k=1}^{K} (\phi^v_{(i, k)} + \phi^w_{(i, k)}) = \phi^{v}_i + \phi^{w}_i$.
In this way, the approximated attribution is proved to satisfy the linearity axiom, \emph{i.e.} $\phi^{v+w}_i = \phi^{v}_i + \phi^{w}_i$.

\textit{Proof of the efficiency axiom:} The approximated attribution of each perturbation pixel is the sum of attributions of corresponding sub-pixels. Thus, the sum of approximated attributions of all perturbation pixels is the sum of attributions of all sub-pixels, \emph{i.e.} $\sum_{i=1}^n \phi_i = \sum_{i=1}^n \sum_{k=1}^K \phi_{(i, k)}$.
 Attributions of sub-pixels satisfy the efficiency axiom, \emph{i.e.} $\sum_{i=1}^n \sum_{k=1}^K \phi_{(i, k)} = z(\Omega^{\text{pixel}}) - z(\emptyset)$. 
 $z(\Omega^{\text{pixel}})$ is the score gained with all sub-pixels, \emph{i.e.} the score made by the whole adversarial perturbation $\delta$, and $z(\emptyset)$ is the score produced without the adversarial perturbation, \emph{i.e.} the score made by the original image.  
 $z(\Omega)$ also represents the score made by the whole adversarial perturbation $\delta$, where $\Omega=\{1, 2, \dots, n\}$ is the set of all perturbation pixels.
 Thus, $z(\Omega^{\text{pixel}}) = z(\Omega)$, and $\sum_{i=1}^n \sum_{k=1}^K \phi_{(i, k)} = z(\Omega^{\text{pixel}}) - z(\emptyset) = z(\Omega) - z(\emptyset)$. 
In this way, the approximated attribution is proved to satisfy the efficiency axiom, \emph{i.e.} $\sum_{i=1}^n \phi_i = z(\Omega) - z(\emptyset)$.

\subsection{Approximation for attributions of sub-pixels based on the Taylor expansion}
\label{appendix:b.3}
In the paper, we approximate attributions of sub-pixels based on the Taylor expansion as Equation~(8). In the supplementary material, we aim to derive the approximation in details. 

Given a function $f(x_1, x_2, \dots, x_n):\mathbb{R}^n\rightarrow\mathbb{R}$, the Taylor expansion at
$(x_1^{(k)}$, $x_2^{(k)}, \dots, x_n^{(k)})$ is 
{
  \begin{equation*}
    \begin{split}
      f(x_1, x_2, \dots, x_n) &= f(x_1^{(k)}, x_2^{(k)}, \dots, x_n^{(k)})  \\
      &+\sum_{i=1}^n (x_i - x_i^{(k)})\frac{\partial f(x_1^{(k)}, x_2^{(k)}, \dots, x_n^{(k)})}{\partial x_i} \\
      &+ {o((x_1-x_1^{(k)}, x_2-x_2^{(k)}, \dots, x_n-x_n^{(k)}))}
    \end{split}
    \end{equation*}
}

$z(S\cup \{(i, k)\})$ denotes the change of the prediction score of the DNN made by sub-pixels in $S\cup \{(i, k)\}$, where $S\subseteq \Omega^{\text{pixel}} \backslash \{(i, k)\}$. The Taylor expansion for $z(S\cup \{(i, k)\})$ at $S$ is given as
{
  \begin{equation*}
    \begin{split}
      z(S\cup \{(i, k)\}) &\approx z(S) + \delta_{(i, k)}\cdot \frac{\partial z(S)}{\partial \delta_{(i, k)}}
    \end{split}
    \end{equation*}
}

Thus, the approximation for the Shapley value of the sub-pixel $(i, k)$ is given as

{
  \begin{equation*}
    \begin{split}
      \phi_{(i, k)}&= \frac{1}{nK}\sum_{S\subseteq \Omega^{\text{pxiel}}\backslash \{ (i, k) \}}\binom{nK-1}{|S|}^{-1}\Big[z(S\cup \{(i, k)\}) - z(S)\Big] \\
      & \approx \frac{1}{nK}\sum_{S\subseteq \Omega^{\text{pixel}}\backslash \{ (i, k) \}}\binom{nK-1}{|S|}^{-1}(\frac{\partial z(S)}{\partial \delta_{(i, k)}}\delta_{(i, k)})
  \end{split}
  \end{equation*}
}

Let there be $m$ components in a certain clustering step. $C^{(u)}_k = \bigcup_{i\in C^{(u)}} (i, k)$ denotes a sub-component. We use $\Omega^{\text{comp}} = \{C^{(1)}_1, C^{(1)}_2, \dots, C^{(m)}_{K-1}, C^{(m)}_K\}$ to denote the set of all sub-components. The Shapley value of the sub-component $C^{(u)}_k$ is approximated as
{
  \begin{equation*}
    \begin{split}
      \phi_{C^{(u)}_{k}}&= \frac{1}{mK}\sum_{S\subseteq \Omega^{\text{comp}}\backslash \{C^{(u)}_{k}\}}\binom{mK-1}{|S|}^{-1}\Big[z(S\cup \{C^{(u)}_k\}) - z(S)\Big]  \\
      &\approx \frac{1}{mK}\sum_{S\subseteq \Omega^{\text{comp}}\backslash \{ C^{(u)}_{k} \}}\binom{mK-1}{|S|}^{-1} \sum_{(i, k)\in C^{(u)}_k}\Big[z(S\cup \{i, k\}) - z(S)\Big] \\
      & \approx \frac{1}{mK}\sum_{S\subseteq \Omega^{\text{comp}}\backslash \{ C^{(u)}_{k} \}}\binom{mK-1}{|S|}^{-1}\sum_{(i, k)\in C^{(u)}_k}(\frac{\partial z(S)}{\partial \delta_{(i, k)}}\delta_{(i, k)}) 
    \end{split}
  \end{equation*}
}

\subsection{Implementation \& computational complexity:}
\label{appendix:b.4}
\begin{framed}
    \textbf{Clarification:} In both the paper and the supplementary material, the computational complexity is quantified as times of network inference, \emph{i.e.} the number of input (masked) images on which we conduct the forward/backward propagation. We do not count the number of detailed operations during the forward/backward propagation \emph{w.r.t.} each specific input image, in order to simplify the analysis. It is because given a specific DNN, the number of detailed operations during the forward/backward propagation is the same for different input images.
\end{framed}

In the paper, we introduce the implementation of the approximation of Shapley values and analyze the computational complexity.
In the supplementary material, we aim to further explain the computational complexity of our approximation for attributions and how we approximate the attribution of components in detail. 

We use a sampling-based method to reduce the complexity of computing Shapley values.
The original formulation of the Shapley value considers all combinations of pixels to compute the Shapley value for each pixel. 
Thus, the computational complexity of the Shapley value of each pixel is $O(2^n)$.
We implement the approximation of Shapley values of sub-pixels with a sampling method.
In this way, the complexity of computing the Shapley value of one sub-pixel is reduced to $O(nKT)$. 
Note that $\phi_i \approx K\cdot \phi_{(i, k)}$. 
Therefore, the complexity of approximating the Shapley value of each pixel is also $O(nKT)$.
The derivatives towards all sub-pixels can be computed simultaneously via back-propagation. 
Thus, the computational complexity of computing Shapley values of all {pixels} remains $O(nKT)$. 

We use hierarchical clustering to iteratively merge several components into a larger component based on interactions. 
We use the following approximation method, to compute and reduce the complexity of computing the attribution of the pair of components.
Let there be $m$ components in a certain clustering step.
Given a component $C^{(u)}$, we can use the sampling method to get their attributions $\phi_{C^{(u)}}$.
Here, from the perspective of game theory, each component is a player, and there are $m$ players in the game.
As mentioned above, the complexity of computing $\phi_{C^{(u)}}$ is $O(mKT)$. 
We use {\small $S^{c} = C^{(i_1)}\cup C^{(i_2)}\cup\dots\cup C^{(i_q)}$} to denote a component candidate.
To determine the interaction inside {\small $S^{c}$}, we need to compute $\phi'_{S^{c}}$. 
The computation of  $\phi'_{S^{c}}$ regards $C^{(i_1)}, C^{(i_2)},\dots, C^{(i_q)}$ as a single component. 
Then, the set of components changes to {\small \{$S^c, C^{(i_{k+1})} \dots, C^{(i_m)}\}$} with $m-q+1$ players. 
In this way, the computational complexity of $\phi'_{S^{c}}$  is {$O((m-q)KT)$}. 
Whereas, considering all potential pairs of components, the computational complexity grows. 
We only consider the interaction between neighboring components.
There are $m$ potential pairs of components, and the complexity of computing all potential pairs of components is {$O(m(m-q)KT)$}.

Considering the local property~\cite{Chen2018L}, we can further approximate {\small $\phi'_{C^{(u)}\bigcup C^{(v)}}$} by simplifying contextual relationships of far-away pixels.
Here, instead of computing {\small  $\phi^\prime_{S^c}$} in the set {\small \{$S^c, C^{(i_{k+1})} \dots, C^{(i_m)}\}$}, we randomly merge $\tilde{m}$ components to get $\tilde{m}/q$ component candidates, including $S^c$.
In this way, the new set includes $\tilde{m}/q$ component candidates and $m-\tilde{m}$ components, \emph{i.e.} {\small $\{S^c, \bigcup_{a=q+1}^{2q}C^{(i_a)}, \dots$, $ \bigcup_{a=\tilde{m}-q+1}^{\tilde{m}}C^{(i_a)}, C^{(i_{\tilde{m}+1})}, \dots, C^{(i_m)}\}$.}
We can simultaneously compute attributions of $\tilde{m}/q$ candidates in the new set, and the computational complexity is {$O((m-(q-1)\tilde{m}/q)KT)$}.
To compute attributions of all potential component candidates, we need to sample $qm / \tilde{m}$ different sets.
In this way, the overall complexity for the computation of attributions of candidates is reduced from {$O(m(m-q)KT)$} to {$O(m(qm/\tilde{m}-q)KT)$}.

\newpage
\section{Pseudo code of extracting perturbation components}
\label{appendix:c}

\begin{algorithm}[h]
    \caption{{Extraction of perturbation components via hierarchical clustering}}
    \label{alg:on-policy}
    \begin{algorithmic}[1]
        \STATE \textbf{Inputs:} pixel set $\Omega$; reward function $z(\cdot)$; component size $q$; iteration times $T$
        \STATE \textbf{Outputs:} Component set $\Omega'$; 
        \STATE \textbf{Initialization:}  $\Omega'=\Omega$
        \FOR{ $iter=1$ to $T$}
        \STATE $\forall C \in \Omega'$, compute $\phi_{C}$ with reward function $z(\cdot)$
        \WHILE{ \textit{not} all possible component candidates are considered}
        \STATE Get component candidate set $\Omega^{\text{candidate}}$ by randomly merging each group of neighboring $q$ components in $\Omega'$
        \STATE $\forall C_{\text{candidate}} \in \Omega^{\text{candidate}}$, compute $\phi_{C_{\text{candidate}}}$ with reward function $z(\cdot)$
        \STATE Compute interaction in each component candidate: $I = \phi_{C_{\text{candidate}}} - \sum_{C\in C_{\text{candidate}}} \phi_{C}$
        \ENDWHILE
        \STATE $\Omega'=\emptyset$
        \STATE Update the component set $\Omega'$ by greedily adding the component candidate with highest interaction strength $|I|$ to $\Omega'$
        \ENDFOR
    \end{algorithmic}
\end{algorithm}

\section{Additional experimental results of regional attributions}
\subsection{Regional attributions computed with different hyper-parameters}
\label{appendix:d.1}
{In this section, we have compared regional attributions with different hyper-parameters ($\beta$ and $L$). The results are shown as follows. We found that important regions indicated by attributions were similar under the same selection of $\beta$, such as the belly region of the pigeon (in the first row) and the wing region of the jaeger (in the second row).  Note that when $\beta$ were different, the generated adversarial perturbations would be different, which leaded to different regional attributions. However, compared with the difference between the magnitudes and attributions, the difference between regional attributions computed with different hyper-parameters was smaller.}

\begin{figure*}[h]

  \centering
  \includegraphics[width=0.9\linewidth]{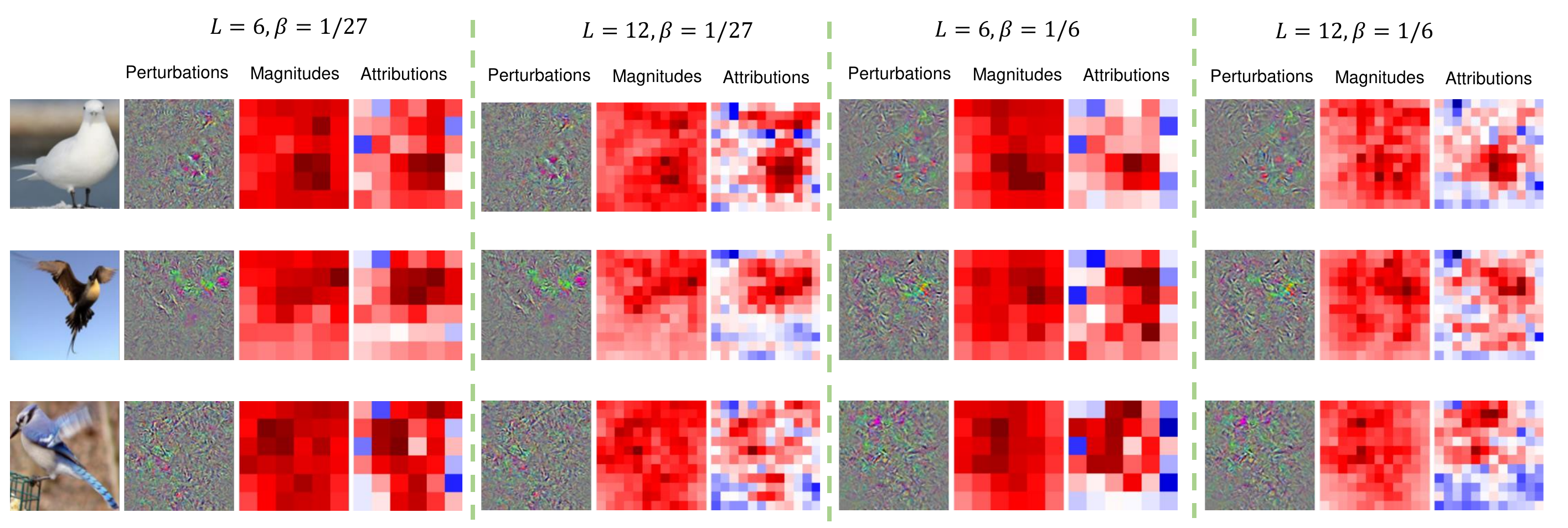}
\end{figure*}

\newpage
\subsection{More experimental results of regional attributions}

Experimental results of regional attributions have been shown in Fig.~{3} in the paper. In the supplementary material, we give additional results of regional attributions.
The visualization shows that although the distribution of $L_2$ adversarial perturbations and the distribution of $L_\infty$ adversarial perturbations were dissimilar, their regional attributions were similar to each other.
\begin{figure*}[h]
  \centering
  \includegraphics[width=0.6\linewidth]{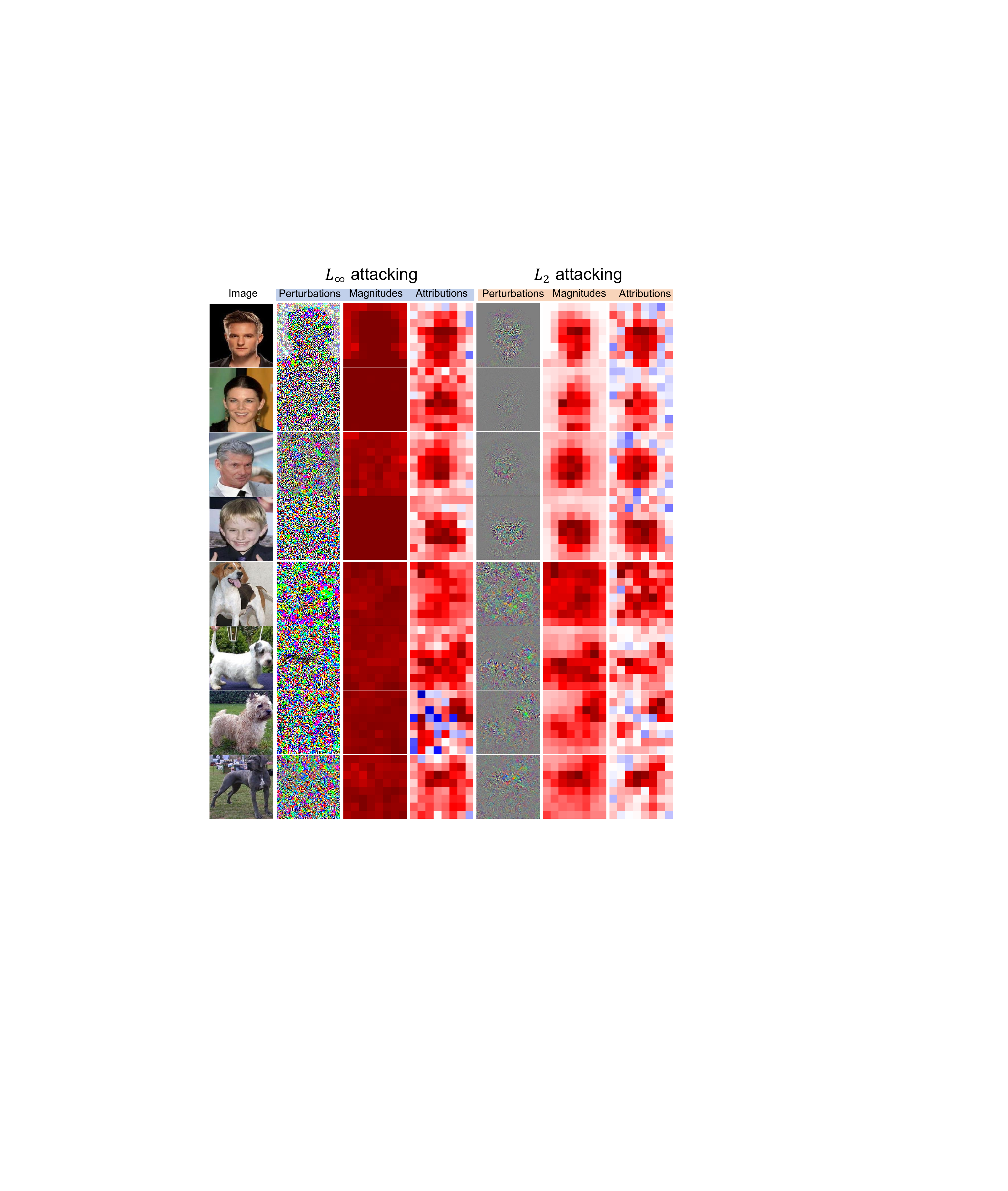}
\end{figure*}

\begin{figure*}[h]
  \centering
  \includegraphics[width=0.6\linewidth]{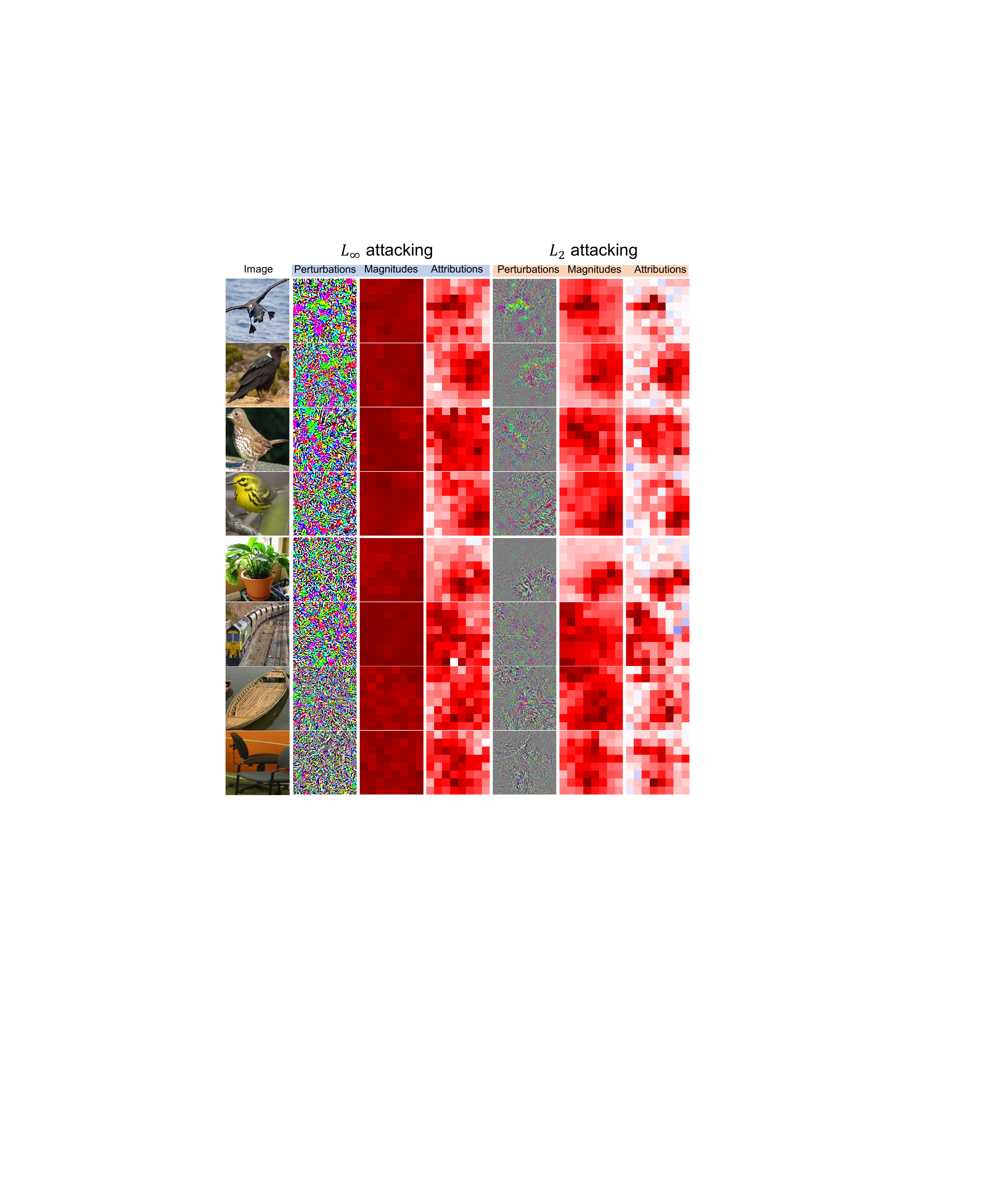}
\end{figure*}

\clearpage
\section{Comparisons of attributions}

There are two types of attributions in the paper, \emph{i.e.} regional attributions to the attacking cost and pixel-level attributions to the change of prediction score (under $L_2$ attacking)).
We visualize regional attributions to the cost of $L_2$ attacking and $L_\infty$ attacking and pixel-wise attribution to the change of the prediction score.
In most cases, important regions indicated by these attributions were similar. For example, in the third row, the dog's head and the horse's body were indicated to be important by all three kinds of attributions. In other cases, important regions indicated by different attributions were different. For example, important regions of the potted plant in the last row indicated by these three kinds of attributions were dissimilar.

\begin{figure}[h!]
  \centering
  \includegraphics[width=0.8\linewidth]{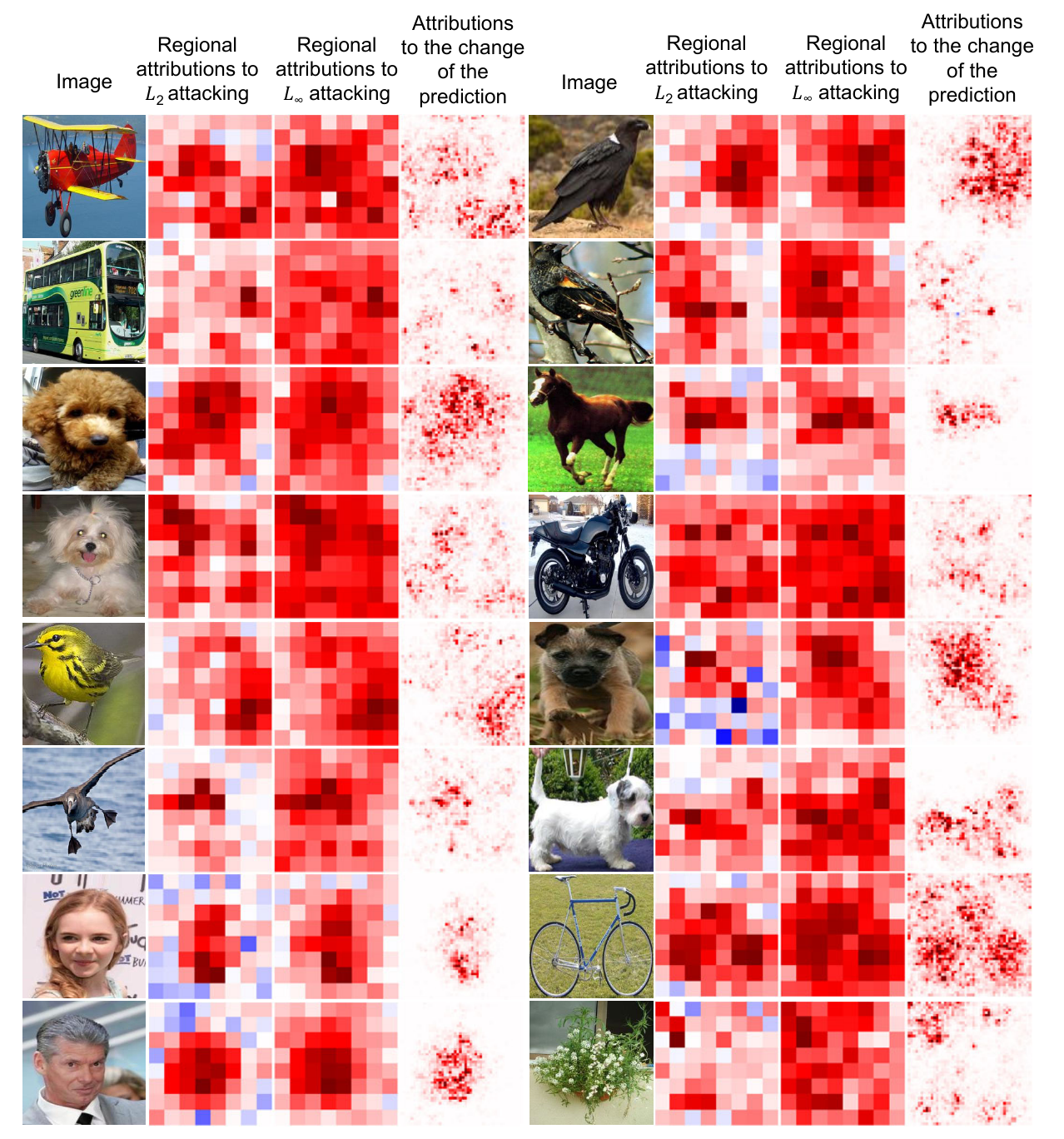}
\end{figure}

\clearpage

\section{More experimental results of interactions and perturbation components}

Experimental results of interactions and perturbation components have been shown in Fig.~4 in the paper. In the supplementary material, we give more examples of visualizations.
Perturbation components usually were not aligned with visual concepts.

\begin{figure}[h!] \label{sup:components}
  \centering
  \includegraphics[width=0.8\linewidth]{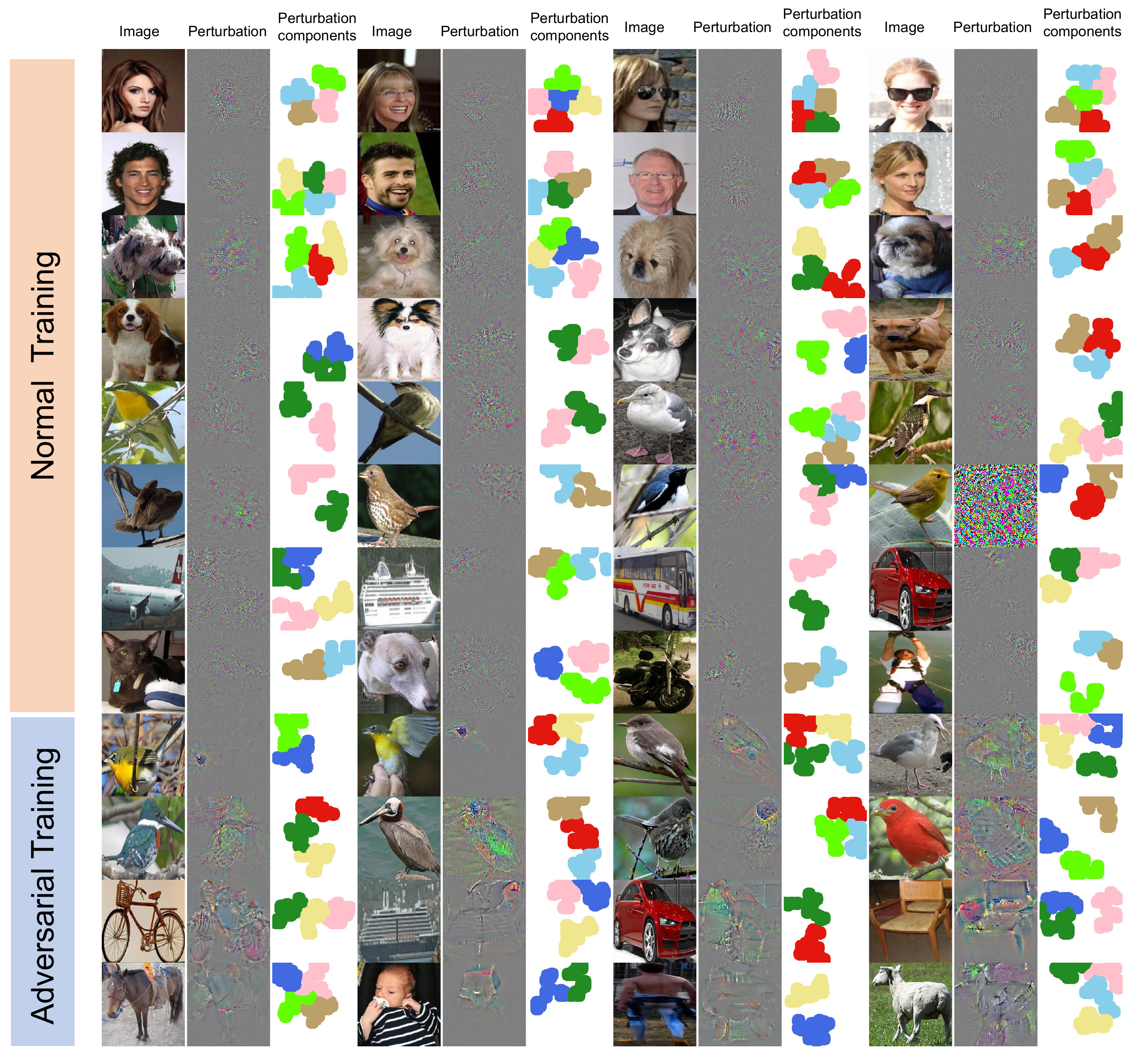}
\end{figure}

\end{document}